\documentclass[runningheads]{llncs}
\usepackage[T1]{fontenc}
\usepackage{cite}
\usepackage{booktabs}
\usepackage{multirow}
\usepackage{xcolor}
\usepackage{graphicx,verbatim}
\begin{document}
\title{Dino U-Net: Exploiting High-Fidelity Dense Features from Foundation Models for Medical Image Segmentation}
%
\author{Haoyue Li\inst{1,2}$^\dagger$ \and
Yifan Gao\inst{1,3}$^\dagger$ \and
Feng Yuan\inst{1,2} \and
Xiaosong Wang\inst{3}$^*$ \and
Jinjiang Cui\inst{2}$^*$ \and
Xin Gao\inst{2,4,5}$^*$}
\authorrunning{H. Li et al.}
\institute{School of Biomedical Engineering (Suzhou), Division of Life Science and Medicine, University of Science and Technology of China, Hefei, China \and
Suzhou Institute of Biomedical Engineering and Technology, Chinese Academy of Sciences, Suzhou, China\and
Shanghai Innovation Institute, Shanghai, China \and
Medical School of Tianjin University, Tianjin, China \and
Jinan Guoke Medical and Technology Development Co., Ltd., Pharmaceutical Valley New Drug Creation Platform, Jinan, China \\
\smallskip
$^\dagger$ These authors contributed equally to this work.\\
$^*$ Corresponding authors.}



\maketitle              
\begin{abstract}
The emergence of large-scale pre-trained visual foundation models offers a promising paradigm for medical image segmentation. However, effectively transferring their rich, generic representations to domain-specific clinical tasks remains a significant challenge. This paper presents Dino U-Net, a novel encoder-decoder framework designed to fully leverage the high-fidelity dense features from the DINOv3 foundation model for medical image segmentation. We employ a frozen DINOv3 backbone as the encoder and introduce a dual-branch DINO Adapter to bridge the feature domain gap. To mitigate the loss of fine-grained information during feature dimensionality reduction, we further propose a Fidelity-Aware Projection Module (FAPM). This module utilizes low-rank shared projection coupled with dynamic feature modulation to refine and faithfully propagate features to the decoder. Extensive experiments on seven public medical image datasets demonstrate that Dino U-Net achieves state-of-the-art performance across diverse imaging modalities, excelling in both regional segmentation accuracy and boundary delineation. Moreover, segmentation performance shows consistent improvement as the backbone scales up to 7 billion parameters, confirming the framework's strong scalability. The code is available at \url{https://github.com/yifangao112/DinoUNet}.


\end{abstract}
\section{Introduction}
Medical image segmentation serves as a cornerstone of computer-aided diagnosis, enabling quantitative analysis for disease screening, treatment planning, and longitudinal monitoring \cite{shaker2024unetr++,dai2021transmed,gao2025wega,gao2023anatomy,gao2025composite}. Consequently, the accuracy and robustness of segmentation algorithms are of paramount importance. Deep learning—particularly the U-Net architecture—has revolutionized this field \cite{unet}. With its efficient encoder-decoder structure and skip connections, U-Net and its variants have become the dominant paradigm, demonstrating consistent performance across diverse imaging modalities \cite{krithika2022review}.  

Notably, foundation models like the Segment Anything Model (SAM) \cite{sam} exhibit impressive zero-shot generalization, with various adaptations for medical tasks \cite{wu2025medical,gao2025safeclick}. However, SAM's reliance on extensive mask-supervision often biases it toward geometric contours, leading to struggles with nuanced semantic relationships or ambiguous anatomical boundaries. In contrast, the self-supervised DINO series \cite{dino,dinov2,dinov3} learns structural patterns and semantic consistency via masked image modeling. By prioritizing semantic density over geometric contours, DINO offers greater robustness for variable-shaped structures. Specifically, the latest DINOv3 mitigates dense feature degradation during large-scale training, providing high-fidelity representations crucial for localizing fine-grained anatomical structures.

In this paper, we propose Dino U-Net, a novel hybrid architecture designed to fully harness the rich, dense features of DINOv3 for medical image segmentation. We construct an encoder using a frozen DINOv3 backbone and introduce a dual-branch DINO Adapter. Furthermore, to preserve fine-grained information during feature dimensionality reduction, we design a Fidelity-Aware Projection Module (FAPM). This module employs low-rank shared projection and dynamic feature modulation to refine features before projection into the decoder. We hypothesize that by leveraging DINOv3’s general high-fidelity representations, U-Net-based segmentation performance can be significantly improved across a wide range of medical imaging tasks.

Our main contributions are as follows: 1) We propose Dino U-Net, a novel architecture that employs a frozen DINOv3 backbone as the encoder integrated with a U-Net decoder. 2) We design a Fidelity-Aware Projection Module (FAPM) to reduce feature dimensionality while maintaining high fidelity. 3) We conduct extensive experiments on seven public datasets, demonstrating state-of-the-art performance and strong generalization across modalities. 4) We show that our method exhibits high scalability, with performance improving as the backbone model scales up.

\section{Related work}
\noindent \textbf{Medical Image Segmentation} The U-Net architecture remains the cornerstone of medical image segmentation. Its encoder-decoder structure and skip connections efficiently capture multi-scale contexts. Numerous variants have optimized this paradigm: U-Net++ \cite{unet++} and SegResNet \cite{segresnet} improve feature propagation, while the nnU-Net \cite{nnunet} framework establishes a robust automated benchmark. Recent models like U-Mamba \cite{umamba} and U-KAN \cite{ukan} further incorporate state-space models and learnable activations to capture long-range dependencies. However, training these models from scratch on limited medical datasets restricts their generalizability across domains. This bottleneck has motivated the transfer of knowledge from large-scale pre-trained foundation models.

\noindent \textbf{Foundation Model-based Medical Image Segmentation} Visual foundation models offer powerful, generalizable feature representations. While prompt-based models like SAM and its successors (SAM2 \cite{sam2}, SAM3 \cite{sam3}) exhibit impressive zero-shot capabilities, their performance on medical images often requires extensive fine-tuning due to significant domain gaps. Parallelly, self-supervised vision transformers, particularly the DINO series \cite{dino, dinov2}, learn robust and semantically dense features. The latest iteration, DINOv3 \cite{dinov3}, provides unprecedented high-fidelity dense representations that rival specialized networks even without fine-tuning. Recently, hybrid architectures like SAM2-UNet \cite{sam2-unet} have successfully integrated foundation encoders into U-Net frameworks. However, the potential of utilizing DINOv3's exceptionally dense features for medical segmentation remains underexplored. To address this gap, we propose Dino U-Net to fully harness DINOv3's high-fidelity representations for elevating segmentation performance.

\subsection{Preliminary}
Vision foundation models, pre-trained on vast datasets of diverse images, have demonstrated a remarkable ability to learn robust and generalizable visual representations. Among these, DINOv3 stands out as a particularly compelling choice for dense prediction tasks such as segmentation. DINOv3 is a self-supervised Vision Transformer, which learns its representations from large-scale, unlabeled natural image data without relying on explicit human annotations. This training paradigm encourages the model to capture a comprehensive understanding of visual scenes, including texture, shape, and context.

The primary advantage of DINOv3 lies in the exceptional quality of its dense features. Unlike its predecessors or other foundation models, DINOv3 was specifically engineered to address the degradation of patch-level feature maps during large-scale training. Through architectural enhancements and a novel training objective named Gram anchoring, it produces feature representations that are remarkably clean and coherent, showing superior performance in tasks that require precise spatial understanding. This proven ability to generate high-fidelity features from general-domain images presents a significant opportunity for medical image segmentation. These rich, dense-pretrained representations can provide a powerful inductive bias for segmenting complex anatomical structures, allowing our model to achieve a new level of accuracy and robustness.

\begin{figure}[ht!]
\centering
\includegraphics[width=\textwidth]{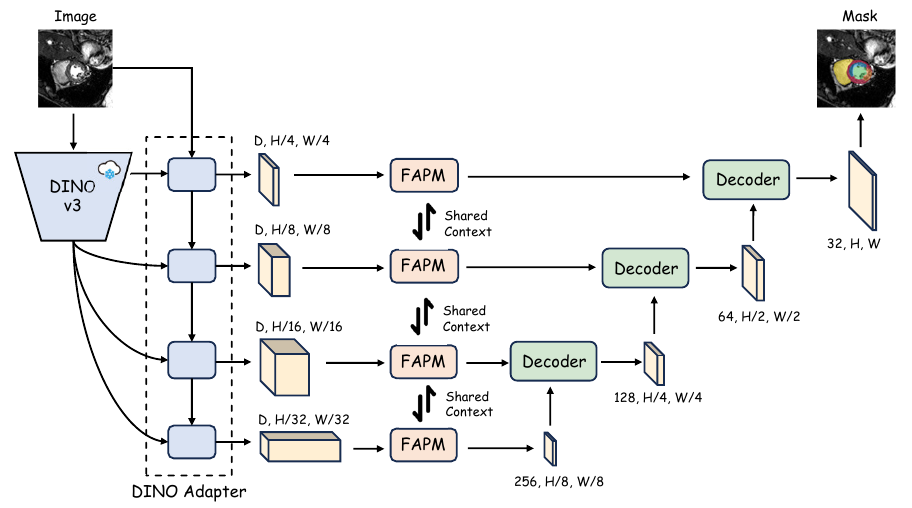}
\caption{Architectural overview of the Dino U-Net. The encoder consists of a frozen DINOv3 backbone, a DINO Adapter, and a Fidelity-Aware Projection Module (FAPM), connected to a standard U-Net decoder.} 
\label{fig:architecture}
\end{figure}
\section{Methodology}

\subsection{Overview of the Dino U-Net}

As illustrated in Fig.~\ref{fig:architecture}, Dino U-Net employs an encoder-decoder design.  The encoder comprises a frozen DINOv3 backbone, a dual-branch DINO Adapter, and a Fidelity-Aware Projection Module (FAPM). 
During forward propagation, the adapter efficiently bridges the domain gap by processing the input concurrently through two branches: a spatial prior branch capturing high-resolution geometric textures, and a semantic branch extracting deep representations from the frozen DINOv3. 
At each interaction stage, a deformable cross-attention mechanism uses geometric cues from the spatial branch to selectively fuse relevant semantic content from DINOv3.
This multi-stage deep fusion yields hierarchical feature maps that perfectly balance spatial acuity with semantic depth. 
Subsequently, these maps are refined by the FAPM to produce high-fidelity representations, which are transmitted via skip connections to the U-Net decoder for progressive spatial recovery and final mask generation. 

\subsection{Fidelity-Aware Projection Module}
\begin{figure}[ht!]
\centering
\includegraphics[width=0.7\linewidth]{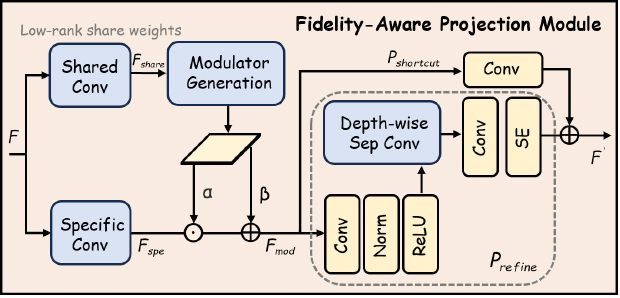}
\caption{Architecture of the FAPM. The input feature maps are first decoupled into task-specific and shared branches. A modulator generation unit utilizes the shared features to dynamically calibrate the specific features. The calibrated features subsequently pass through a refinement path ($P_{refine}$) and are merged with a shortcut path ($P_{shortcut}$) to generate the high-fidelity output.}
\label{fig:fapm_architecture}
\end{figure}

The DINO Adapter produces semantically rich multi-scale features ($F$) in the high-dimensional DINOv3 space, which are incompatible with the channel dimensions of a standard U-Net decoder. Naïve projection methods (e.g., linear layers or 1×1 convolutions) often sacrifice fine-grained details, compromising feature fidelity. To address this, we propose the Fidelity-Aware Projection Module (FAPM) for dimensionality reduction while preserving feature integrity (see Fig.~\ref{fig:fapm_architecture}). 

FAPM first decouples input feature $F$ via a dual-branch structure: a Shared Conv branch with low-rank shared weights (channel dimension 256) extracts cross-scale contextual information $F_{share}$, while a Specific Conv branch captures scale-specific spatial details $F_{spe}$. Then, $F_{share}$ is fed into the Modulator Generation unit (a lightweight $1 \times 1$ convolution) to produce two spatially adaptive parameters: the scaling factor $\alpha$ and the shifting factor $\beta$. These parameters perform an affine transformation on the specific feature $F_{spe}$. Specifically, $F_{spe}$ is multiplied by $\alpha$ and then added to $\beta$. This process enhances local details, yielding the modulated feature $F_{mod}$:
\begin{equation}
    F_{mod} = F_{spe} \odot \alpha  + \beta
    \label{eq:fapm_mod}
\end{equation}
where $\odot$ denotes element-wise multiplication.

$F_{mod}$ is processed through dual paths for integration: a refinement path $P_{refine}$ employing depthwise separable convolutions and a Squeeze-and-Excitation (SE) block to polish local details, and a shortcut path $P_{shortcut}$ that uses a convolution to align channels and preserve the original feature manifold. The final high-fidelity output $F'$ is obtained as:
\begin{equation}
    F' = P_{refine}(F_{mod}) + P_{shortcut}(F_{mod})
    \label{eq:fapm_output}
\end{equation}
which is subsequently passed via skip connection to the U-Net decoder to facilitate high-resolution mask reconstruction.

\section{Experiments}
\textbf{Datasets} To comprehensively evaluate the effectiveness and generalization capabilities of the proposed Dino U-Net, we conducted extensive experiments on seven public datasets. 
These datasets were selected to cover diverse challenges in medical image segmentation, encompassing multiple imaging modalities (e.g., MRI, ultrasound, endoscopy), various anatomical targets, and different pathology types. Detailed characteristics of each dataset are listed in Table \ref{dataset}.

\begin{table*}[]
\centering
\caption{Summary of the seven public datasets used for evaluation.}
\label{dataset}
\resizebox{\textwidth}{!}{
\begin{tabular}{ccccc}
\toprule
\textbf{Dataset} & \textbf{Category}                 & \textbf{Modality} & \textbf{Target Classes}                  & \textbf{Cases} \\ \midrule
Kvasir-SEG \cite{jha2020kvasir}       & Polyp                & Endoscopy         & Polyps                                   & 1000           \\

BUSI \cite{al2020dataset}         & Breast Tumor                & Ultrasound        & Benign \& Malignant Tumors               & 780            \\
CellBinDB \cite{shi2025cellbindb}      & Cell                 & Microscopy        & 
Cells (Mouse \& Human)                   & 1044           \\
PROSTATEx-Seg-Zones \cite{meyer2020prostatex}  & Gland                & MRI               & 4 Prostate Zones                         & 98             \\
Drishti-GS \cite{6867807}       & Optic Nerve Head     & Fundus            & Optic Disc \& Cup                        & 101            \\
MyoPS20 \cite{gao2023bayeseg,zhuang2018multivariate,qiu2023myops}      & Myocardial Pathology & CMR               & Scar, Edema, Myocardium, LV/RV Pools     & 25             \\
m2caiSeg \cite{maqbool2020m2caiseg}   & Surgical Instrument         & Endoscopy         & 19 classes (Organs, Instruments, Fluids) & 307            \\ \bottomrule
\end{tabular}
}
\end{table*}

\begin{table*}[ht!]
\centering
\caption{Comparison of segmentation performance on seven medical datasets. We report the Dice (\%) and the HD95. For the Dice, higher is better ($\uparrow$); for HD95, lower is better ($\downarrow$). For our Dino U-Net variants, scores outperforming the best baseline are marked in \textbf{bold}, and scores outperforming the second-best baseline are \underline{underlined}.}
\label{tab:segmentation_results_revised}
\resizebox{\textwidth}{!}{%
\begin{tabular}{l|cc|cc|cc|cc|cc|cc|cc}
\toprule
\multicolumn{1}{l|}{\multirow{2}{*}{\textbf{Method}}} & \multicolumn{2}{c|}{\textbf{Kvasir-SEG}} & \multicolumn{2}{c|}{\textbf{Drishti-GS}} & \multicolumn{2}{c|}{\textbf{BUSI}} & \multicolumn{2}{c|}{\textbf{CellBinDB}} & \multicolumn{2}{c|}{\textbf{MyoPS20}} & \multicolumn{2}{c|}{\textbf{PROSTATEx}} & \multicolumn{2}{c}{\textbf{m2caiseg}} \\
& Dice$\uparrow$ & HD95$\downarrow$ & Dice$\uparrow$ & HD95$\downarrow$ & Dice$\uparrow$ & HD95$\downarrow$ & Dice$\uparrow$ & HD95$\downarrow$ & Dice$\uparrow$ & HD95$\downarrow$ & Dice$\uparrow$ & HD95$\downarrow$ & Dice$\uparrow$ & HD95$\downarrow$ \\
\midrule
nnU-Net \cite{nnunet} & 80.85 & 61.48 & 82.94 & 1.60 & 65.51 & 77.06 & \underline{87.82} & 2.71 & 61.98 & 22.46 & 69.52 & 5.11 & 51.55 & 50.48 \\
SegResNet \cite{segresnet} & 85.98 & 35.08 & 82.91 & 1.62 & 67.83 & 64.95 & 87.03 & 3.12 & 67.60 & 14.24 & 71.93 & 4.62 & 43.18 & 53.72 \\
UNet++ \cite{unet++} & 85.69 & 44.08 & 82.82 & 1.65 & 68.96 & 66.63 & 87.43 & 2.89 & 67.42 & 16.33 & 71.82 & 4.76 & 49.80 & \underline{43.03} \\
U-Mamba \cite{umamba} & 85.24 & 40.91 & 82.16 & 1.78 & 68.53 & 56.05 & 87.74 & 3.15 & 62.51 & 28.00 & 72.02 & 4.89 & 52.49 & 45.74 \\
U-KAN \cite{ukan} & 85.33 & 41.32 & 83.89 & 1.48 & 68.29 & \textbf{44.88} & 86.99 & 3.92 & 73.59 & 11.15 & \underline{72.65} & 4.54 & 51.01 & 45.28 \\
Swin U-Mamba\dag \cite{swinumamba} & 81.16 & 59.59 & 80.06 & 2.03 & 66.43 & 66.20 & 86.90 & 3.15 & 70.39 & 11.00 & 69.81 & 5.57 & 44.65 & 51.94 \\
SAM2-UNet \cite{sam2-unet} & \underline{89.79} & \underline{27.07} & 83.43 & 1.64 & 66.04 & 59.20 & 87.23 & 3.22 & 74.07 & 10.39 & 71.67 & 4.39 & 49.70 & 52.74 \\
\midrule
Dino U-Net S & 88.87 & 33.24 & 84.19 & \underline{1.43}$^{\sim}$ & \textbf{71.87}$^{\ddag}$ & 58.40 & 87.38 & 2.80 & \underline{74.92}$^{\ddag}$ & 10.21$^{\ddag}$ & 72.02 & \textbf{4.16}$^{\sim}$ & 51.13 & 52.08 \\
Dino U-Net B & 88.56 & 28.03 & 83.90 & 1.47 & 70.78$^{\ddag}$ & 57.36 & 87.56 & 3.19 & \textbf{75.41}$^{\ddag}$ & \underline{8.39}$^{\ddag}$ & 72.19 & 4.38 & 52.39 & 47.23 \\
Dino U-Net L & 89.63 & 27.81 & \underline{84.41} & \textbf{1.40}$^{\sim}$ & \underline{71.25}$^{\ddag}$ & 56.44 & 87.51 & \textbf{2.53}$^{\sim}$ & 74.22$^{\sim}$ & 10.49 & 72.50 & \underline{4.33} & \underline{52.78}$^{\sim}$ & 45.53 \\
Dino U-Net 7B & \textbf{90.77}$^{\ddag}$ & \textbf{25.44}$^{\ddag}$ & \textbf{84.48}$^{\sim}$ & 1.48 & 70.63$^{\ddag}$ & \underline{46.85} & \textbf{87.92}$^{\sim}$ & \underline{2.64} & 74.63$^{\ddag}$ & \textbf{8.35}$^{\ddag}$ & \textbf{73.09}$^{\sim}$ & 4.53 & \textbf{53.46}$^{\ddag}$ & \textbf{42.01}$^{\ddag}$ \\
\bottomrule
\multicolumn{15}{l}{$\sim$:p-value>0.05, $\ddag$:p-value<0.05, in comparison between the best and best baseline results.} \\
\end{tabular}
}
\end{table*}

\begin{table}[ht!]
\centering
\caption{Overall efficiency and average performance. Efficiency is measured by the number of active parameters (Params). \textbf{Bold} denotes the best results; \textcolor{red}{red} indicates improvement over the best baseline.}
\label{tab:performance_efficiency_delta}
\resizebox{0.55\textwidth}{!}{
\begin{tabular}{l ccc}
\toprule
\textbf{Method} & \textbf{Params} $\downarrow$ & \textbf{Dice (\%)} $\uparrow$ & \textbf{HD95} $\downarrow$ \\
\midrule
nnU-Net \cite{nnunet}     & 30.45 & 71.45 & 31.56 \\
SegResNet \cite{segresnet} & 10.26 & 72.35 & 25.34 \\
UNet++ \cite{unet++}       & 14.39 & 73.42 & 25.62 \\
U-Mamba \cite{umamba}      & 63.76 & 72.96 & 25.79 \\
U-KAN \cite{ukan}          & 9.38  & 74.54 & 21.80 \\
Swin U-Mamba$^\dagger$ \cite{swinumamba} & 37.76 & 71.34 & 28.50 \\
SAM2-UNet \cite{sam2-unet} & 5.38  & 74.56 & 22.66 \\
\midrule
Dino U-Net S  & \textbf{5.11} & 75.77 \textcolor{red}{(+1.21)} & 23.19 \\
Dino U-Net B  & 11.65 & 75.83 \textcolor{red}{(+1.27)} & 21.44 \textcolor{red}{(-0.36)} \\
Dino U-Net L  & 18.14 & 76.04 \textcolor{red}{(+1.48)} & 21.22 \textcolor{red}{(-0.58)} \\
Dino U-Net 7B & 228.97 & \textbf{76.43} \textcolor{red}{(+1.87)} & \textbf{18.76} \textcolor{red}{(-3.04)} \\
\bottomrule
\end{tabular}}
\end{table}

\noindent \textbf{Comparison Methods and Metrics} We evaluated four Dino U-Net variants (S, B, L, and 7B, scaled by backbone size) against seven state-of-the-art baselines: CNN-based architectures (nnU-Net, SegResNet, UNet++), recent Mamba/KAN-based models (U-Mamba, U-KAN), and foundation-model adaptations (Swin U-Mamba$^\dagger$ \cite{swinumamba}, SAM2-UNet). Segmentation accuracy was measured using the Dice Similarity Coefficient (Dice) and 95\% Hausdorff Distance (HD95). Additionally, we reported the number of active parameters to evaluate model efficiency and scalability. We assessed statistical significance using the Wilcoxon signed-rank test ($p < 0.05$).

\begin{figure*}[ht!]
\centering
\includegraphics[width=\textwidth]{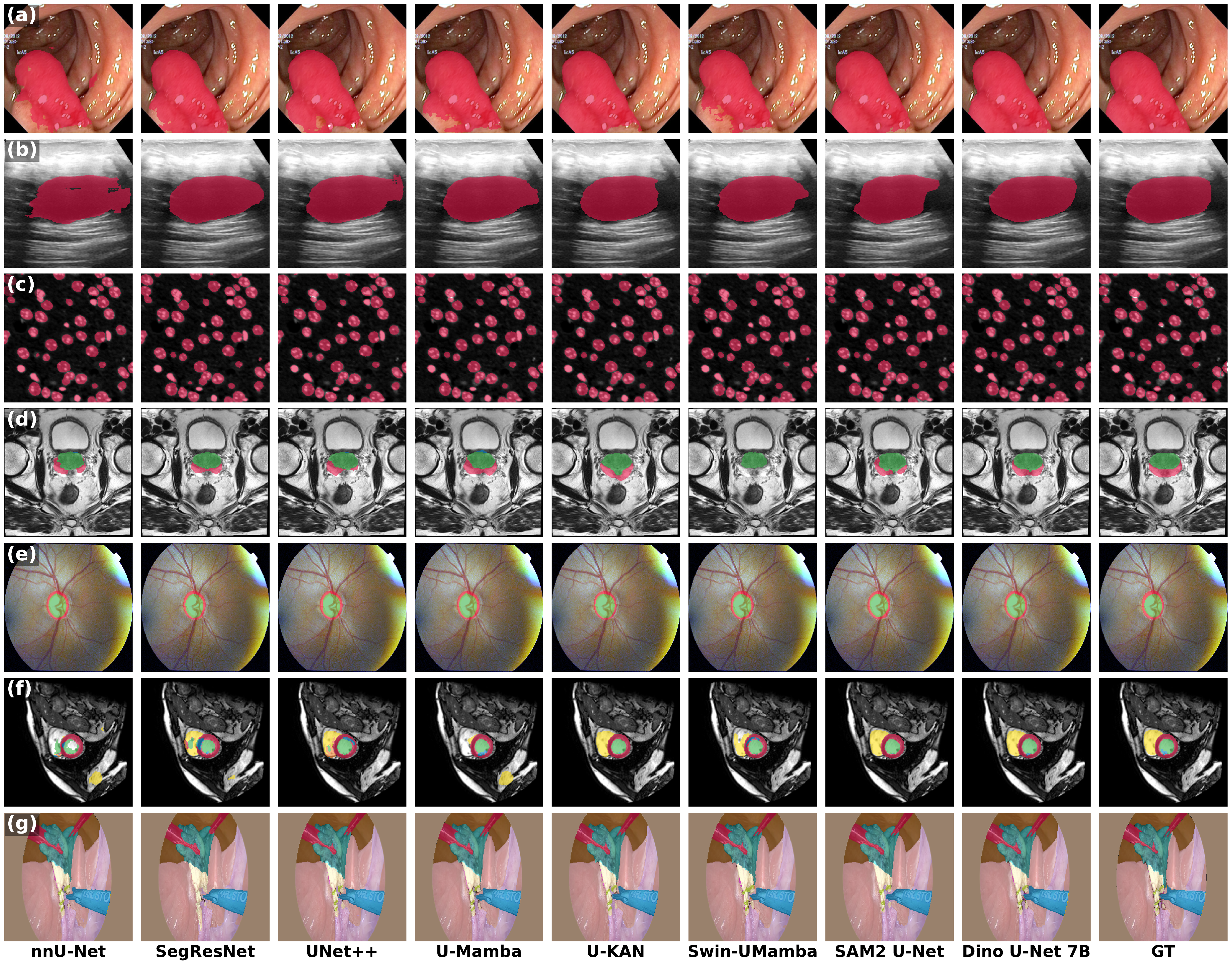}
\caption{Qualitative comparison of segmentation results on representative samples from the seven evaluation datasets. Each row corresponds to a different dataset (from top to bottom: Kvasir-SEG, BUSI, CellBinDB, PROSTATEx-Seg-Zones, Drishti-GS, MyoPS20, and m2caiSeg). Columns display the results of different comparison methods, our method (Dino U-Net 7B), and the Ground Truth (GT).}
\label{fig:segmentation_results}
\end{figure*}

\noindent \textbf{Implementation details}
Implemented in PyTorch, we strictly followed the default nnU-Net pipeline for preprocessing, data augmentation, and sliding-window inference. Datasets were randomly split into 8:2 training and testing sets. Models were trained using a combined Dice and Cross-Entropy loss, optimized by Adam (initial learning rate
$1 \times 10^{-3}$ with polynomial decay). Training lasted 200 epochs (250 iterations per epoch) on NVIDIA H100 GPUs.

\begin{table}[htbp]
    \centering
    \caption{Ablation study on the FAPM. We report the relative changes in parameter counts ($\Delta$ Params), Dice score ($\Delta$ Dice), and HD95 ($\Delta$ HD95) after integrating the FAPM module across various model scales. $\uparrow$ indicates higher is better, and $\downarrow$ indicates lower is better.}
    \label{tab:fapm_ablation}
    \begin{tabular*}{0.5\textwidth}{@{\extracolsep{\fill}}lcccc@{\extracolsep{\fill}}}
        \toprule
        \multirow{2}{*}{\textbf{Metrics}} & \multicolumn{4}{c}{\textbf{Model Size}} \\
        \cmidrule{2-5}
        & \textbf{S} & \textbf{B} & \textbf{L} & \textbf{7B} \\
        \midrule
        $\Delta$ Params (M)             & $+0.5$  & $+0.3$  & $+0.2$  & $-0.8$  \\
        $\Delta$ Dice (\%) $\uparrow$   & $+0.79$ & $+0.77$ & $+0.56$ & $+0.75$ \\
        $\Delta$ HD95 (mm) $\downarrow$ & $-1.75$ & $-0.09$ & $-0.37$ & $-1.04$ \\
        \bottomrule
    \end{tabular*}
\end{table}

\noindent \textbf{Results}
Quantitative evaluations (Table \ref{tab:segmentation_results_revised} and Table \ref{tab:performance_efficiency_delta}) demonstrate that Dino U-Net consistently outperforms seven state-of-the-art methods across diverse modalities, with improvements being statistically significant in most cases. Notably, the largest 7B variant sets new records on five of the seven datasets, achieving the best average Dice score of 76.43\% (+1.87\%) and HD95 of 18.76 (-3.04). The architecture also exhibits exceptional scalability and efficiency: performance improves consistently as the backbone scales, yet even the lightweight small variant surpasses the average performance of all baselines. Qualitatively (Fig.~\ref{fig:segmentation_results}), our method achieves more precise boundary delineation than contour-reliant models like SAM2-UNet, showing superior robustness in low-contrast textures.

\noindent \textbf{Ablation Study}
To validate FAPM, we replaced it with simple 1×1 convolutions across all scales (Table~\ref{tab:fapm_ablation}). Regarding efficiency, FAPM's low-rank shared basis design demonstrates scalability: while adding marginal parameters to smaller models (S, B, L), it effectively reduces the overall parameter count for the largest 7B model. Furthermore, removing FAPM consistently degrades segmentation accuracy across all scales, with Dice dropping by up to 0.79\% and HD95 worsening by up to 1.75mm. This confirms that FAPM is crucial for preserving high-fidelity features and precise boundary delineation.
\section{Discussion and Conclusion}
In this paper, we propose Dino U-Net, which integrates the DINOv3 foundation model with a dual-branch adapter and a novel FAPM to leverage high-fidelity dense features from large-scale pre-training. Dino U-Net’s superior performance stems from leveraging DINOv3’s self-supervised representations and the FAPM’s high-fidelity projection, which provide a more robust semantic inductive bias than contour-biased models. The scaling success from small version to 7B variants further confirms that medical segmentation significantly benefits from the massive representational power of billion-parameter foundation models. Despite these strengths, the current framework is limited to 2D tasks, necessitating future extension to 3D volumetric data. Additionally, while the frozen backbone ensures parameter efficiency, the inference overhead of the 7B model remains high, suggesting future research into knowledge distillation to develop more compact yet powerful models.

Extensive experiments on seven public datasets confirm that our method significantly outperforms state-of-the-art models across various imaging modalities. This study demonstrates that the rich, pre-trained representations from next-generation self-supervised models provide a superior inductive bias for medical segmentation. Future work will focus on extending this framework to 3D volumetric segmentation tasks.
%
%
    

\begin{credits}
\subsubsection{\ackname} This work was supported by the National Science Foundation of China (Grant No. 82372052, U24A20757, 82371957), and the Suzhou Science and Technology Plan Project (Grant No. SZS2025008, SYG2024141).

\subsubsection{\discintname}
The authors declare that they have no conflicts of interest.
\end{credits}

%
%
%
\bibliographystyle{splncs04}
\bibliography{reference.bib}
%




\end{document}